\documentclass[lettersize,journal]{IEEEtran}
\usepackage{amsmath,amsfonts}
\usepackage{array}
\usepackage[caption=false,font=normalsize,labelfont=sf,textfont=sf]{subfig}
\usepackage{textcomp}
\usepackage{stfloats}
\usepackage{url}
\usepackage{verbatim}
\usepackage{graphicx}
\usepackage{cite}

\usepackage[utf8]{inputenc} 
\usepackage[T1]{fontenc}    
\usepackage{hyperref}       
\usepackage{url}            
\usepackage{booktabs}       
\usepackage{amsfonts}       
\usepackage{nicefrac}       
\usepackage{microtype}      
\usepackage{xcolor}         

\usepackage[utf8]{inputenc} 
\usepackage[T1]{fontenc}    
\usepackage{hyperref}       
\usepackage{url}            
\usepackage{booktabs}       
\usepackage{amsfonts}       
\usepackage{nicefrac}       
\usepackage{microtype}      
\usepackage{lipsum}
\usepackage{fancyhdr}       
\usepackage{graphicx}       
\graphicspath{{media/}}     
\usepackage{mathtools}
\usepackage{amsmath,amssymb,amsfonts}
\usepackage{algorithmic}

\usepackage{graphicx}
\usepackage{textcomp}
\usepackage{xcolor}
\usepackage{ulem}
\usepackage{comment}
\usepackage[ruled,vlined]{algorithm2e}
\usepackage[T1]{fontenc}
\usepackage[utf8]{inputenc}
\usepackage{verbatim}
\usepackage{soul} 
\usepackage[hyphenbreaks]{breakurl}
 \usepackage{float}
\usepackage{hyperref}
\usepackage[hyphenbreaks]{breakurl}
\usepackage{multirow}
\usepackage{adjustbox}

\usepackage{ulem}

\usepackage{enumerate}
\usepackage{colortbl}
\usepackage{booktabs}
\usepackage{array}
\usepackage[english]{babel}
\usepackage{amsthm}
\usepackage{setspace}
\theoremstyle{plain}
\newtheorem{theorem}{Theorem}

\theoremstyle{definition}

\theoremstyle{remark}
\newtheorem{remark}[theorem]{Remark}

\begin{document}

\title{Learning on Bandwidth Constrained Multi-Source Data with MIMO-inspired DPP MAP Inference}

\author{Xiwen Chen, Huayu Li, Rahul Amin, Abolfazl Razi~\IEEEmembership{Senior Member,~IEEE}


        
}

\markboth{Journal of \LaTeX\ Class Files,~Vol.~14, No.~8, August~2021}%
{Shell \MakeLowercase{\textit{et al.}}: A Sample Article Using IEEEtran.cls for IEEE Journals}


\maketitle

\begin{abstract}

This paper proposes a distributed version of Determinant Point Processing (DPP) inference to enhance multi-source data diversification under limited communication bandwidth. DPP is a popular probabilistic approach that improves data diversity by enforcing the repulsion of elements in the selected subsets. The well-studied Maximum A Posteriori (MAP) inference in DPP aims to identify the subset with the highest diversity quantified by DPP. However, this approach is limited by the presumption that all data samples are available at one point, which hinders its applicability to real-world applications such as traffic datasets where data samples are distributed across sources and communication between them is band-limited.

Inspired by the techniques used in \textit{Multiple-Input Multiple-Output} (MIMO) communication systems, we propose a strategy for performing MAP inference among distributed sources. Specifically, we show that a lower bound of the diversity-maximized distributed sample selection problem can be decomposed into a sum of MIMO-like sub-problems. 
A determinant-preserved sparse representation of selected samples is used to perform sample \textit{Pre-coding} in local sources to be processed by DPP. 
Our method does not require raw data exchange among sources, but rather a band-limited feedback channel to send lightweight diversity measures, analogous to the \textit{Channel State Information} (CSI) message in MIMO systems, from the center to data sources.
The experiments show that our scalable approach can outperform alternative methods and finally demonstrates the potential to translate the Diversification to the improvement of learning quality in several applications, such as multi-level classification, object detection, and multiple-instance learning.
\end{abstract}


\section{Introduction}
\begin{figure}[!htbp]
\centering\includegraphics[width=0.4\textwidth]{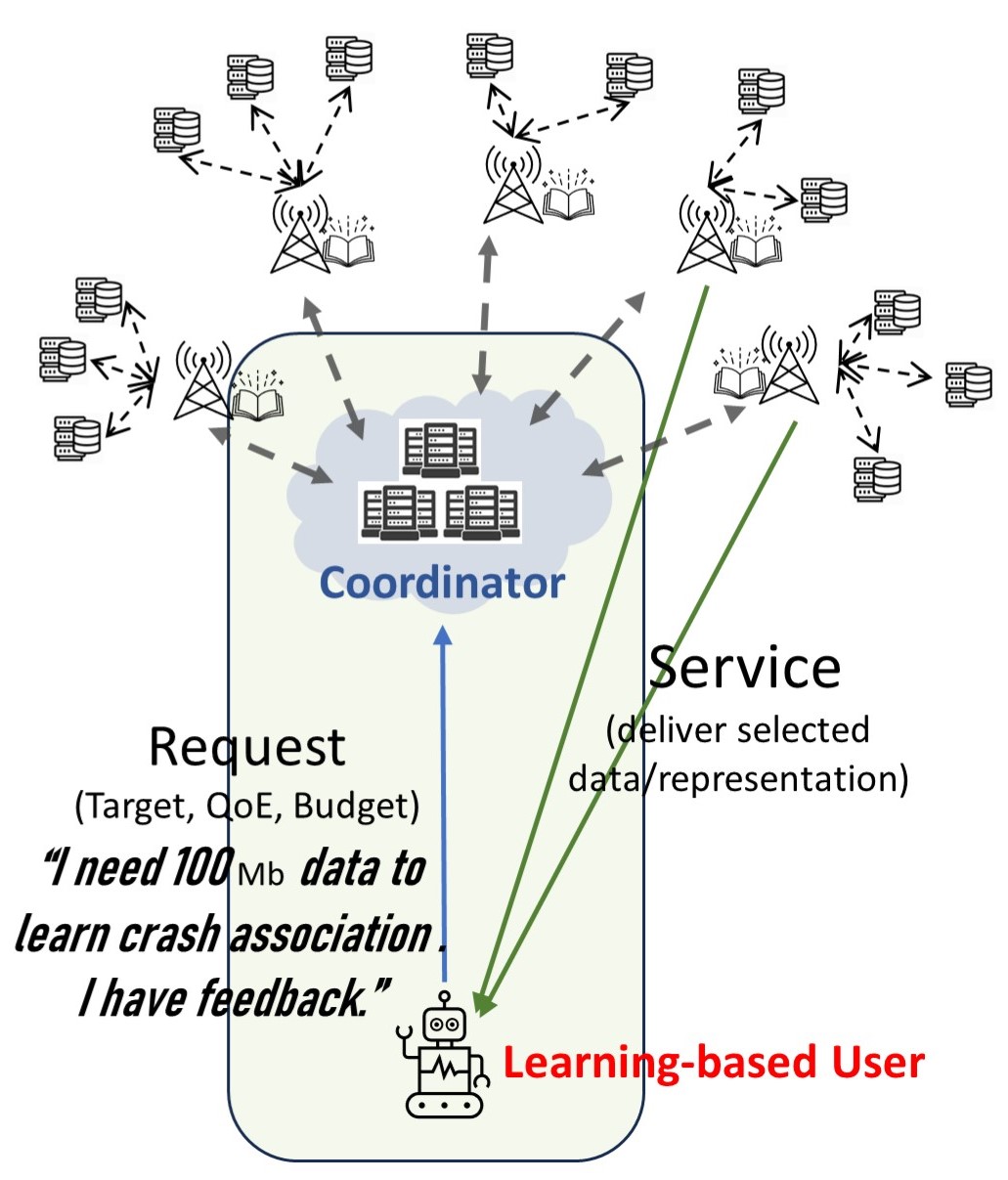}
\caption{The workflow of the data delivery system.  A user requests data; the Coordinator executes distributed diverse sampling and schedules Edge servers to deliver data. The allocation is dynamic based on the feedback from the user. For simplicity, in this work, we describe "center" as the role of both the coordinator and the user. } 
\label{fig:s1}
\end{figure}

In recent years, with the explosive growth of data, using a centralized source to store and process data collected from various sensors has become inefficient and even infeasible because of the capacity of storage and communication limitations. Therefore, collecting and storing the data in distributed sources and then delivering data to users when they request become a valuable trend and has been widely studied in AI-enabled semantic communication networks \cite{shi2021semantic}. It is noteworthy that in many learning-based platforms (shown in Fig. \ref{fig:s1}), the ultimate goal is to improve the \textit{Quality of Experience} (QoE) to the users by extracting semantic information from the delivered data, not just the data itself. Therefore, the strategy to deliver the most useful data becomes a critical problem, especially when the transmission between sources and users is constrained. 
Adopting random selection strategies is relieving but not optimal in most cases. 
Indeed, it has been known for decades that \textit{diversity} of selected samples can dramatically enhance the quality of learning applications \cite{ayinde2019regularizing,wu2021entropy,yu2022can,zhou2023inverse}. Hence, selecting data samples that best mimic the geometrical distribution of the entire dataset by maximizing cross-sample distances in the original or transformed domain is explicit for this purpose. Due to its simple form, high interpretability, and high efficiency,
\textit{Determinantal point processes} (DPP), is commonly used for diversity maximization.

DPP is a well-known probabilistic approach to facilitate generating diverse data points.
DPP stands out among other point processing techniques, such as Poisson point processing, for its unique property of being solely determined by the correlation among elements. It assigns a high probability to the measurement of two points with low similarity, making it a valuable tool for tasks like dimensionality reduction and representative sample selection from large datasets \cite{kulesza2012determinantal}. Additionally, by leveraging the properties of linear algebra, DPP is able to sample subsets from a given data set in an efficient manner \cite{derezinski2019exact,calandriello2020sampling,derezinski2019fast}.  
DPP is used in a wide range of applications, such as recommender systems \cite{chen2018fast}, document summarization \cite{perez2021multi}, image processing \cite{launay2021determinantal}, and topological analysis of wireless networking \cite{deng2014ginibre}. It also inherently connects to different principles, such as randomized numerical linear algebra \cite{derezinski2021determinantal}, and information theory \cite{chen2023rd}. 

As mentioned earlier, in order to use data for learning tasks, we expect to identify and select the most diverse subset. This goal implemented by DPP is often known as the \textit{Maximum A Posteriori} (MAP) inference problem. 
Recent studies have implemented a centralized version of this algorithm, where all samples are available in the same location \cite{chen2018fast,han2020map,bhaskara2020online}. 
However, in the applications of semantic communication networks, data samples are collected by different sources at different positions, and cross-source communication is challenging or costly. In our scenario, the communication bandwidth between data sources and the processing center is constrained. Although we assume that the selected items can be transmitted losslessly to the center, but the transmission budget allows only sending a subset of the entire dataset. Therefore, a conventional DPP MAP inference \cite{chen2018fast} by traversing all data samples is infeasible. 
In the context, the most popular method \cite{mirzasoleiman2016distributed} to perform distributed DPP inference is multi-stage, which basically implements a local greedy search by each source to collect candidate samples regardless of other sources’ samples and then performs another selection on the accumulated candidates to obtain the final samples. 
However, a zero-communication overhead policy by merely sending the selected samples is enforced in our scenario, whereas this popular method, involves transmitting extra samples that are ultimately discarded by the center. Therefore, its performance may be challenged in our scenario.

\begin{figure*}[!ht]
\centering\includegraphics[width=0.8\textwidth]{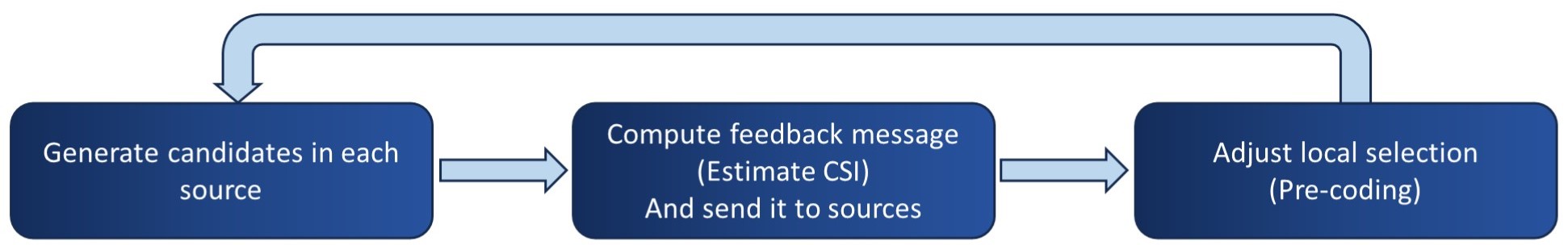}
\caption{The workflow of the proposed MIMO-like selection framework.} 
\label{fig:block}
\end{figure*}

In this paper, inspired by the procedures of Multiple-Input Multiple-Output (MIMO) systems, which are commonly used in contemporary wireless communication systems, we propose an efficient and scaleable scheduling strategy to select data with a feedback mechanism (shown in Fig. \ref{fig:block}). Specifically, the problem is analogous to simplifying the evaluation of power-optimized MIMO capacity, and 
the transmission candidates are generated locally without exchanging raw samples among sources. When the transmission is in an interval-by-interval fashion, a spare representation capturing the effect of currently accumulated samples is transmitted through a lightweight feedback channel from the center to the sources to adjust the local selecting strategy. These two steps are analogous to the estimation of \textit{Channel State Information} (CSI) estimation and \textit{Pre-coding}.

\textbf{Contributions.} In summary, we propose a MIMO-inspired strategy for performing the MAP inference on distributed data under limited communication constraints with the following steps. First, we reformulate the lower bound of the diversity, which is used to decompose the joint optimization into a sum of individual problems. Afterward, we show the solution can be improved by the feedback message via diversifying the conditional lower bound. 
Then, we give a determinant-preserving approximation of the feedback message based on Cauchy–Binet's formula to address the bandwidth-limited transmission. Finally, we demonstrate data selection by our method in proper representation can enhance the learning quality in multiple applications, such as classification, object detection, and weakly-supervised multiple-instance learning (MIL). 

\section{Background Knowledge}
\subsection{Determinant Point Processing (DPP)}
DPP is a probability measure on all $2^{|\mathcal{S}|}$ subsets of $\mathcal{S}$, where $|\mathcal{S}|$ denotes the cardinality of the set $\mathcal{S}$. Suppose a finite dataset is represented by $\mathbf{Z}=[\mathbf{z}_1^\top,\mathbf{z}_2^\top,\cdots,\mathbf{z}_n^\top]\in \mathbb{R}^{n\times m}$. 
Given a \textit{Gram Matrix} $\mathbf{L}=\mathbf{Z} \mathbf{Z}^{\top}$, \textit{L-ensemble DPP} is presented by having an arbitrary subset $A$ drawn from the entire set $\mathcal{S}$ to satisfy,
$
\mathcal{P}({A})\propto\operatorname{det}\left(\mathbf{L}_{{A}}\right),
$
where $\mathbf{L}_{{A}}$ denotes the submatrix of $\mathbf{L}$ with rows and columns indexed by set $A$. The MAP inference for K-DPP is formulated as,
\begin{align}
\arg \max_{A\subseteq S}~\operatorname{det} (\mathbf{L}_{{A}}),\quad s.t.|A| = k_T,
\setstretch{0.6}
\end{align}%
where $A$ denotes the index set of selected samples and constant $k_T$ denotes the given fixed cardinality, and $k_T\leq rank(\mathbf{L})$ holds to ensure the determinant greater than 0 \cite{horn2012matrix}. Since MAP inference is an NP-hard problem, the popular solution is using greedy search and formulating the following sub-modular function,
$
 j=\arg \max _{i \in S \backslash A} \log \operatorname{det}\left(\mathbf{L}_{A \cup\{i\}}\right)-\log \operatorname{det}\left(\mathbf{L}_{A}\right)$, which can give a $(1-1/e)$-approximation of the optimal.
Here, $j$ denotes the selected index in each round.
The current fastest greedy search proposed in \cite{chen2018fast} is based on the Cholesky decomposition and requires $\mathcal{O}(n^3)$ complexity for initialization and $\mathcal{O}(k_T^2n)$ to return $k_T$ items. We denote selection by this method with given Gram matrix $\mathbf{L}$ and the set cardinality $k_T$ as $A^* = \text{MAP-DPP}(\mathbf{L},k_T)$. 
\subsection{Multiple-Input Multiple-Output (MIMO) systems}
Before delving into the design of our selection scheduling for DPP inference on distributed data, let us briefly review the fundamental principles of MIMO systems, due to their techniques serve as valuable inspiration for our proposed approach.
In a MIMO system, a signal vector $\mathbf{s}\in\mathbb{C}^{m_t}$ is transmitted by $M$ antennas ($TX_1, \cdots, TX_{m_t}$, ) to be received as a vector $\mathbf{r}\in\mathbb{C}^{{m_r}}$ by $N$ antennas ($RX_1, \cdots, RX_{m_r}$). The link between a transmitting antenna $TX_i$ and a receiving antenna $RX_j$ is represented by an element  $\mathbf{G}_{i,j}$ of the channel matrix $\mathbf{G} \in \mathbb{C}^{{m_t} \times {m_r}}$. The channels are influenced by factors such as multi-path fading and interference, causing different link conditions. Mathematically, a MIMO system can be presented as $\mathbf{r} = \mathbf{G}\mathbf{s}+\mathbf{n}$,
where $\mathbf{n}\in\mathbb{C}^{{m_r}}$ is additive noise that follows a complex Gaussian distribution $\mathcal{CN}(0,\sigma^2\mathbf{I})$ with zero mean and covariance $\sigma^2\mathbf{I}$. Apparently, when the channels are highly correlated and the rank of $\mathbf{G}$ is low, the equations for recovering $\mathbf{s}$ from $\mathbf{r}$ become under-determined. To mitigate the interference of different antennas at the receiver, \textit{pre-coding} is employed to orthogonalize data between channels by utilizing channel state information (CSI).

We assume $m_t=m_r$ and let $\mathbf{Q}=\mathbb{E}[\mathbf{s}\mathbf{s}^\dagger]$ denote the covariance matrix of $\mathbf{s}$, where $\mathbf{s}^\dagger$ denotes the conjugate transpose of $\mathbf{s}$. Inequality $\text{trace}(\mathbf{Q})\leq \rho$ always holds for preserving the overall power constraint. The capacity of the system measures the maximum amount of information that can be transferred with an arbitrary small error. According to~\cite{gesbert2003theory}, the capacity is:
\begin{align}\label{eq:mimo}
\setlength{\abovedisplayskip}{3pt}
\setlength{\belowdisplayskip}{3pt}
  \max_{\mathbf{Q}}  C{=}\log_2 \operatorname{det}(\mathbf{I}+\frac{1}{\sigma^2}\mathbf{G}\mathbf{Q}\mathbf{G}^\dagger)~~~s.t.~\text{trace}(\mathbf{Q})\leq \rho.
\end{align}
The SVD of $\mathbf{G}$ is denoted by $\mathbf{U}\mathbf{S}\mathbf{V}^\dagger= svd(\mathbf{G})$, and $\lambda_i$ represents the $i$-th singular value of $\mathbf{G}$, which corresponds to the diagonal element of $\mathbf{S}$. Consequentially, the optimal solution of $\mathbf{Q}$ can be expressed as $\mathbf{V}\mathbf{P}\mathbf{V}^\dagger$, where $\mathbf{P}$ is a diagonal matrix with elements $p_i$. Then, the problem in Eq. (\ref{eq:mimo}) can be reformulated accordingly:
\begin{align}\label{eq:mimo2}
    &\max _{\mathbf{P}: \sum_{i=1}^N p_i \leq \rho} \log_2 \operatorname{det}\left(\mathbf{I}+\frac{1}{\sigma^2} \mathbf{U} \mathbf{S}^2 \mathbf{P} \mathbf{U}^\dagger\right) \\ \nonumber
    \overset{(a)}{=}&\max _{\mathbf{P}: \sum_{i=1}^N p_i \leq \rho} \sum_{i=1}^N \log \left(1+\frac{p_i \lambda_i^2}{\sigma^2}\right).
\end{align}
Noting, $\log \left(1+\frac{p_i \lambda_i^2}{\sigma^2}\right)$ is concave and indicates the capacity of the single-input single-output (SISO) channel. The original problem can be transformed by (a) in Eq. (\ref{eq:mimo2}) and solved by standard optimization algorithms. 
Additionally, the SVD-based precoding scheme can be applied to pre-code the signal to $\widetilde{\mathbf{r}}= \mathbf{V}\mathbf{P}^{1/2}\mathbf{r}$, which helps orthogonalize the data.

\section{Problem Formulation}\label{sec:Formulation}
It is noteworthy that notations from this section are shown in Table \ref{tab:notations}, which may slightly change from previous sections.
Suppose there are $N$ data sources with disjointed index sets $S_1,S_2, \cdots,S_i,\cdots,S_N$, and $\mathcal{S}=S_1\cup S_2\cup S_i\cup\cdots\cup S_N$ representing the indices of the entire set, and $S_i\cap S_j=\emptyset,\forall i\neq j$. The total number of samples is $n=\sum_{i=1}^N n_i$, where $n_i = |S_i|$ denotes the cardinality of the set $S_i$. Also, let $\mathbf{Z}\in \mathbb{R}^{n\times m}$ be the data matrix of the entire set, where the dimensions of each sample $\mathbf{z}_i$ is $m$. Recall that to maximize diversity, we need to optimize the following problem under communication constraints to select a subset $\mathcal{A}$:
\begin{align}\label{eq:1}
       \arg \max_{\mathcal{A}\subseteq \mathcal{S}} ~\operatorname{det} (\mathbf{L}_\mathcal{A}), \quad s.t. &|\mathcal{A}| = k_T, 
       \vspace{-0.1in}
    \end{align}
where, again, $\mathbf{L}_\mathcal{A}$ denotes the columns and rows of $\mathbf{L}$ indexed by $\mathcal{A}$. Likewise, $X_{A,B}$ denotes a submatrix of $X$ with rows and columns indexed by $A$ and $B$, respectively. If $A=B$ and $X$ is a square, it can be denoted as $X_A$, otherwise $X_A$ only denotes the rows of a non-squared matrix $X$ indexed by $A$ (i.e. $\mathbf{Z}_{\mathcal{A}}$).
$k_T$ denotes the total of samples that can be selected under bandwidth limitation.
Conventional MAP methods require transmitting all samples to the center to solve this optimization problem, which is impractical or costly when data is distributed across different sources. Therefore, the construction of $\mathbf{L}$ is not realistic. However, $\mathbf{L}_{S_1}, \cdots,\mathbf{L}_{S_N}$ can be easily obtained at different sources, where $\mathbf{L}_{S_i}= \mathbf{Z}_{S_i} \mathbf{Z}_{S_i}^\top$, and  the subproblems of global MAP inference 
\begin{align}\label{eq:2}
\arg \max_{A_i\subseteq S_i} ~\operatorname{det} \left(\left(\mathbf{L}_{S_i}\right)_{A_i}\right), \quad s.t. |A_i| = k_i
\end{align}
 can be solved locally. Here, $\left(\mathbf{L}_{S_i}\right)_{A_i}$ denotes the matrix $\mathbf{L}_{S_i}$ indexed by $A_i$. The problem now is to collectively select $A_i$ for source $i$, which can be achieved by maximizing of $\operatorname{det} (\mathbf{L}_\mathcal{A})$ with $\mathcal{A}=A_1\cup A_2\cup A_i\cup\cdots\cup A_N $ subject to the constraint $\sum|A_i|=k_T$. Following \cite{mirzasoleiman2016distributed}, we assume each source transmits the same number of samples, which is denoted as, $k_i=k_T/N$.
\vspace{-0.1in}

\section{Methodology}\label{sec:Methodology}
 \subsection{MIMO-like Decomposition}
Similar to \cite{chen2023rd,yang2008unsupervised}, since $\mathbf{L}_\mathcal{A}$ is always a positive-definite Hermitian matrix, we can use the approximation $\operatorname{det} (\frac{1}{\epsilon}\mathbf{L}_{\mathcal{A}})\approx \operatorname{det} (\frac{1}{\epsilon}\mathbf{L}_\mathcal{A}+\mathbf{I})$ for a very small $\epsilon$. Therefore, we can rewrite the optimization problem as,
\begin{align}\label{eq:new_problem}
 \setstretch{0.6}
    \arg\max_\mathcal{A} \log\operatorname{det} (\frac{1}{\epsilon}\mathbf{L}_\mathcal{A}+\mathbf{I}) &=\log \operatorname{det} (\frac{1}{\epsilon}\mathbf{Z}_\mathcal{A}\mathbf{Z}_\mathcal{A}^\top+\mathbf{I}) \\   \nonumber
    &\overset{(a)}{=}\log \operatorname{det} (\frac{1}{\epsilon}\mathbf{Z}_\mathcal{A}^\top\mathbf{Z}_\mathcal{A}+\mathbf{I}) \\ \nonumber
    &=
    \log \operatorname{det} (\frac{1}{\epsilon}\sum_i^N\mathbf{Z}_{A_i}^\top\mathbf{Z}_{A_i}+\mathbf{I}).\nonumber
     \setstretch{0.6}
\end{align}
The validity of  (a) in Eq. (\ref{eq:new_problem}) can be established through SVD decomposition. When $N\ll \frac{1}{\epsilon}$ holds, the maximization problem can be approximated as 
\begin{align}
\arg\max\mathcal{L}= \log\operatorname{det} (\frac{1}{\epsilon}\sum_i^N\mathbf{Z}_{A_i}^\top\mathbf{Z}_{A_i}/N+\mathbf{I}).
\end{align}
\begin{theorem}
    The lower bound of the approximated problem in Eq. (\ref{eq:new_problem}) is given by $\mathcal{L}^{lower}=\frac{1}{N}\sum_i^N \log \operatorname{det} (\frac{1}{\epsilon}\mathbf{Z}_{A_i}\mathbf{Z}_{A_i}^\top+\mathbf{I})$.
    
\end{theorem}
\begin{proof}
    First, we apply the concavity of $f(A)=\log \operatorname{det} A$ for positive definite Hermitian matrices \cite{horn2012matrix}, we have, $\log \operatorname{det}(\alpha A+(1-\alpha) B) \geq \alpha \log \operatorname{det} A+(1-\alpha) \log \operatorname{det} B$ for $\alpha\in(0,1)$. 
     Therefore, we can easily obtain, $\mathcal{L}\geq\mathcal{L}^{lower} = \frac{1}{N}\sum_i^N \log \operatorname{det} (\frac{1}{\epsilon}\mathbf{Z}_{A_i}^\top\mathbf{Z}_{A_i}+\mathbf{I})$.
    Then we apply the equality presented in Eq. (\ref{eq:new_problem})(a), the proof is completed. 
\end{proof}
    
\begin{remark}
    theorem 1 allows us to decompose the global optimization problem into the sum of sub-problems, which is analogous to the procedure of MIMO systems.
\end{remark}

\subsection{MIMO-like CSI Estimation and Pre-coding}
Similar to the MIMO systems, we aim here to alleviate the interference of different sources. Ideally, we expect that whatever the selection from other sources, each source should perform the selection process in parallel, and then simply collect the samples from all sources can achieve the maximized global diversity. The solution is to make the block-diagonalization of the similarity matrix $\mathbf{L}$, which can be analogous to orthogonalization in MIMO.
Decoupling the correlations of the sources helps minimize the gap between the approximated and original problems. We can achieve this goal implicitly by \textit{pre-coding} the samples in each source $\mathbf{Z}_{A_i}$ to $\widetilde{\mathbf{Z}}_{A_i}$ by $\widetilde{\mathbf{Z}}_{A_i}=\mathbf{Z}_{A_i}\mathbf{W}_i$. However, learning $\mathbf{W}_i$ by accessing extra data samples in all sources may not be feasible due to the limited communication budget. Instead, we propose a sparse diversity measurement of selected samples that serves as the pre-coding matrix to tighten the lower bound and guide the selection process.

Recall that $\mathcal{A}$ represents the collection of all selected samples after the completion of the entire selection process. We then denote selected samples until a moment by $\mathcal{B}\subseteq \mathcal{A}$ and define $Y_i$ as the index set of selected samples at this moment that are \textbf{not} from source $S_i$, i.e., $Y_i= \mathcal{B}\setminus A_i$. It is noteworthy that at that moment, $\mathcal{A}$ is not fully known.

In each source, to avoid losing information on samples selected from the current local source (because there will be information loss if we compress the diversity measurement of selected samples in the center), we decided to use selection with replacement. This approach ensures the selected samples will not be dropped in the source after sending. We define $\mathcal{Y}=\{Y_1,\cdots,Y_N\}$, where we also have $\mathcal{B}=Y_1\cup\cdots \cup Y_N$. Therefore, we can rewrite the approximation of the problem in Eq. (\ref{eq:new_problem}) when conditioned by $\mathcal{Y}$ as
\begin{align}\label{eq:cond_problem}
\setlength{\abovedisplayskip}{3pt}
 \mathcal{L}_{\text{real}}&=\log \operatorname{det} (\frac{1}{\epsilon}\sum_i^N\mathbf{Z}_{A_i}^\top\mathbf{Z}_{A_i}+\mathbf{I})\\ \nonumber
 &\overset{(a)}{\geq} \mathcal{L}(\cdot\mid \mathcal{Y})\\ \nonumber
 &\overset{(b)}{:=}\log\operatorname{det} (\frac{1}{\epsilon}\sum_i^N(\mathbf{Z}_{A_i}^\top\mathbf{Z}_{A_i}+\mathbf{Z}_{Y_i}^\top\mathbf{Z}_{Y_i})/N +\mathbf{I}). 
\end{align}
where equality (a) holds when $\mathcal{A}=A_i\cup Y_i=\mathcal{B}$ and (b) defines the conditional problem. The conditional lower bound $\mathcal{L}(\cdot\mid \mathcal{Y})$ can be obtained as,
\begin{align}\label{eq:lowerband2}
   \mathcal{L}(\cdot\mid \mathcal{Y})&\overset{(a)}{\geq}  \mathcal{L}^{lower}(\cdot\mid \mathcal{Y})\\ \nonumber &\overset{}{:=} \frac{1}{N}\sum_i^N \log \operatorname{det} (\frac{1}{\epsilon}\mathbf{Z}_{A_i}^\top\mathbf{Z}_{A_i}+\frac{1}{\epsilon}\mathbf{Z}_{Y_i}^\top\mathbf{Z}_{Y_i}+\mathbf{I}) \\ \nonumber &\overset{}{=} \frac{1}{N}\sum_i^N \log \operatorname{det} (\frac{1}{\epsilon}\mathbf{Z}_{{\{A_i\cup Y_i}\}}\mathbf{Z}_{{\{A_i\cup Y_i}\}}^\top+\mathbf{I}),
\end{align}
where equality (a) holds when $\mathcal{A}=A_i\cup Y_i=\mathcal{B}$.
\begin{theorem}
$\mathcal{L}^{lower}(\cdot\mid \mathcal{Y})$ is a tighter bound than $\mathcal{L}^{lower}$ (presented in Theorem 1) to the problem $\mathcal{L}_{\text{real}}$, which is presented as 
$
 \mathcal{L}_{\text{real}}\overset{}{\geq}\mathcal{L}^{lower}(\cdot\mid \mathcal{Y})\overset{}{\geq}\mathcal{L}^{lower}
 $.
\end{theorem}
\begin{proof}
    To prove 
    \begin{align}
&\log \operatorname{det} (\frac{1}{\epsilon}\sum_i^N\mathbf{Z}_{A_i}^\top\mathbf{Z}_{A_i}+\mathbf{I})\\ \nonumber
&\overset{(a)}{\geq}\log\operatorname{det} (\frac{1}{\epsilon}\sum_i^N(\mathbf{Z}_{A_i}^\top\mathbf{Z}_{A_i}+\mathbf{Z}_{Y_i}^\top\mathbf{Z}_{Y_i})/N +\mathbf{I}),
    \end{align}
we can convert $LFS$ and $RHS$ to a form of submodular function. According to Eq. (\ref{eq:new_problem}), $LFS=\log\operatorname{det} (\frac{1}{\epsilon}\mathbf{L}_\mathcal{A}+\mathbf{I})$.\\
Applying the fact  $A_i\cup Y_i=\mathcal{B}$, which implies $\mathbf{Z}_{A_i}^\top\mathbf{Z}_{A_i}+\mathbf{Z}_{Y_i}^\top\mathbf{Z}_{Y_i}=\mathbf{Z}_{\mathcal{B}}^\top\mathbf{Z}_{\mathcal{B}}$, therefore,
$
RHS=
\log\operatorname{det} (\frac{1}{\epsilon}\sum_i^N(\mathbf{Z}_{\mathcal{B}}^\top\mathbf{Z}_{\mathcal{B}})/N +\mathbf{I})=\log\operatorname{det} (\frac{1}{\epsilon}\mathbf{Z}_{\mathcal{B}}\mathbf{Z}_{\mathcal{B}}^\top +\mathbf{I})=\log\operatorname{det} (\frac{1}{\epsilon}\mathbf{L}_\mathcal{B}+\mathbf{I}) ,
$
Now, $LFS$ and $RHS$ is the same form of a submodular function w.r.t $A$ and $\mathcal{B}$, respectively. Since $\mathcal{B}\subseteq A$, the inequality is proved.
\end{proof}

\begin{remark}\label{remark:4}
    We note that $\mathcal{L}^{lower}(\cdot\mid \mathcal{Y})$ is still a form of the sum of subproblems. Therefore, this theorem demonstrates that, by taking the received information $\mathcal{Y}$ merely from the center as feedback, we can adjust the future selection in each source to enhance the overall diversity. 
\end{remark}
\begin{remark}
in fact, $\mathcal{L}^{lower}(\cdot\mid \mathcal{Y})$ can be updated sequentially as$
    \mathcal{L}^{lower}(\cdot\mid \mathcal{Y}): \mathcal{L}^{lower}(\cdot\mid \mathcal{Y}^0)\rightarrow \mathcal{L}^{lower}(\cdot\mid \mathcal{Y}^1)\rightarrow \cdots,
$ if the selection is online.
Here, $\mathcal{Y}^t$ is formed after receiving data samples in the first $t$ intervals.
\end{remark}

Now, according to Remark \ref{remark:4}, the target of each source is switched to perform local selection adjusted by feedback, which is equivalent to maximize $\log \operatorname{det} (\frac{1}{\epsilon}\mathbf{Z}_{{\{A_i\cup Y_i}\}}\mathbf{Z}_{{\{A_i\cup Y_i}\}}^\top+\mathbf{I})$ individually in each source. Fortunately, this problem rolls back to a MAP inference as  
\begin{align}\label{eq:}
    &\max_{A_i\subseteq S_i} ~\operatorname{det} \left(\mathbf{Z}_{{\{A_i\cup Y_i}\}}\mathbf{Z}_{{\{A_i\cup Y_i}\}}^\top\right),\\ \nonumber    &s.t. \quad|A_i| = k_i,~A_i\cap Y_i = \emptyset.
\end{align}
According to Schur's complement, we can re-write it as,
\begin{align}\label{eq:12}
   \max_{A_i\subseteq S_i}~ &\operatorname{det} (\mathbf{Z}_{{\{A_i\cup Y_i}\}}\mathbf{Z}_{{\{A_i\cup Y_i}\}}^\top) 
   \\ \nonumber  
      =&\operatorname{det}( \mathbf{Z}_{Y_i} \mathbf{Z}_{Y_i}^\top)\operatorname{det}\left(\mathbf{Z}_{A_i} \left(\mathbf{I}- \mathbf{Z}_{Y_i}^\top ( \mathbf{Z}_{Y_i} \mathbf{Z}_{Y_i}^\top)^{-1} \mathbf{Z}_{Y_i}\right)  \mathbf{Z}_{A_i}^\top\right).
\end{align}
Since $\operatorname{det}( \mathbf{Z}_{Y_i} \mathbf{Z}_{Y_i}^\top)$ is fixed, the only information depending on $Y_i$ is $\mathbf{H}_i:=\left(\mathbf{I}- \mathbf{Z}_{Y_i}^\top ( \mathbf{Z}_{Y_i} \mathbf{Z}_{Y_i}^\top)^{-1} \mathbf{Z}_{Y_i}\right)$. Without loss of information, we can send $\mathbf{H}_i$ (which can be viewed as the CSI) from the center to each source and adjust (which can be viewed as the pre-coding) $\mathbf{Z}_{A_i}$ to $\widetilde{\mathbf{Z}}_{A_i} = \mathbf{Z}_{A_i} \mathbf{H}_i^{1/2}$, where $\mathbf{W}_i = \mathbf{H}_i^{1/2}$. Now, we only need to maximize $\operatorname{det} (\mathbf{Z}_{A_i} \mathbf{H}_i^{1/2}\mathbf{H}_i^{1/2}\mathbf{Z}_{A_i}^\top)$.
\subsection{Sparse Representation of MIMO-like CSI}\label{sec:csi}
To further accommodate the band-limited communication requirements, we seek sending a sparse representation of $\mathbf{H}_i$ than sending the entire information of $\mathbf{H}_i$. To this end, we use the Cauchy–Binet formula to re-write the optimization problem as,
\begin{align}\label{eq:all}
    \max_{A_i}~~&\operatorname{det} (\mathbf{Z}_{A_i} \mathbf{H}_i^{1/2}\mathbf{H}_i^{1/2}\mathbf{Z}_{A_i}^\top) \\ \nonumber
    =& \operatorname{det} (( \mathbf{Z}_{A_i} \mathbf{H}_i^{1/2})( \mathbf{Z}_{A_i} \mathbf{H}_i^{1/2})^\top)  \\ \nonumber
    =& \sum_{J_1 \in\left(\begin{array}{c}
{[m]} \\
k_i
\end{array}\right)} \left(\operatorname{det}\left(( \mathbf{Z}_{A_i} \mathbf{H}_i^{1/2})_{[k_i], J_1}\right)\right)^2,\\ \nonumber
\end{align} 
where $[m]$ denotes the set $  \{1, ..., m\}$ and $\left(\begin{array}{c}{[m]} \\k_i\end{array}\right)$ denotes the set of $k_i$-combinations of $[m]$. As before, $k_i$ denotes the cardinality of $A_i$.

\begin{theorem}
    The lower bound of $\left(\operatorname{det}\left(( \mathbf{Z}_{A_i} \mathbf{H}_i^{1/2})_{[k_i], J_1}\right)\right)^2$ is given by $\operatorname{det}(\mathbf{Z}_{A_i}\mathbf{Z}_{A_i}^\top)\operatorname{det}(\mathbf{H}_{J_1})$, which is presented as
    \begin{align}\label{eq:part}
        \left(\operatorname{det}\left(( \mathbf{Z}_{A_i} \mathbf{H}_i^{1/2})_{[k_i], J_1}\right)\right)^2\leq \operatorname{det}(\mathbf{Z}_{A_i}\mathbf{Z}_{A_i}^\top)\operatorname{det}(\mathbf{H}_{J_1}).
    \end{align}
\end{theorem}

\begin{proof}
    \begin{align}
&\left(\operatorname{det}\left(( \mathbf{Z}_{A_i} \mathbf{H}_i^{1/2})_{[k_i], J_1}\right)\right)^2 
=\left(\operatorname{det}\left(\mathbf{Z}_{A_i}(\mathbf{H}_i^{1/2})_{[m], J_1}\right)\right)^2 \\ \nonumber
=&\left(\sum_{J_2 \in\left(\begin{array}{c}
{[m]} \\
k_i
\end{array}\right)} \operatorname{det}\left((\mathbf{Z}_{A_i})_{[k_i], J_2}\right) \operatorname{det}\left((\mathbf{H}_i^{1/2})_{J_2,J_1}\right)\right)^2 \\ \nonumber
\overset{(a)}{\leq}& \sum_{J_2 \in\left(\begin{array}{c}
{[m]} \\
k_i
\end{array}\right)} \left(\operatorname{det}\left((\mathbf{Z}_{A_i})_{[k_i], J_2}\right)\right)^2 \\ \nonumber &\times\sum_{J_2 \in\left(\begin{array}{c}
{[m]} \\
k_i
\end{array}\right)} \left(\operatorname{det}\left((\mathbf{H}_i^{1/2})_{J_2,J_1}\right)\right)^2 \\ \nonumber
\overset{(b)}{=}&\sum_{J_2 \in\left(\begin{array}{c}
{[m]} \\
k_i
\end{array}\right)} \left(\operatorname{det}\left((\mathbf{Z}_{A_i})_{[k_i], J_2}\right)\right)^2 \\ \nonumber &\times\operatorname{det}\left((\mathbf{H}_i^{1/2})_{J_1,[m]}((\mathbf{H}_i^{1/2})_{J_1,[m]})^\top\right) \\ \nonumber
\overset{(c)}{=}& \sum_{J_2 \in\left(\begin{array}{c}
{[m]} \\
k_i
\end{array}\right)} \left(\operatorname{det}\left((\mathbf{Z}_{A_i})_{[k_i], J_2}\right)\right)^2 \operatorname{det}(\mathbf{H}_{J_1})\\ \nonumber 
\overset{(d)}{=}& \operatorname{det}(\mathbf{Z}_{A_i}\mathbf{Z}_{A_i}^\top)\operatorname{det}(\mathbf{H}_{J_1}),
 \setstretch{0.6}
\end{align} 
where (a) applies Cauchy–Schwarz inequality and (b)(c)(d) apply Cauchy–Binet formula.
\end{proof}

 Now we consider $\hat{\mathbf{H}}_i$ denotes the approximate $\mathbf{H}_i$ by its sparse representation. Substituting Eq. (\ref{eq:part}) to  Eq. (\ref{eq:all}) to preserve the determinant of $( \mathbf{Z}_{A_i} \hat{\mathbf{H}}_i^{1/2}\hat{\mathbf{H}}_i^{1/2} \mathbf{Z}_{A_i}^\top) $, we should ensure 
\begin{align}\label{eq:ensure}
\sum_{J_1 \in\left(\begin{array}{c}
{[m]} \\
k_i
\end{array}\right)}\operatorname{det}((\hat{\mathbf{H}}_i)_{J_1})\overset{(a)}{\approx}\sum_{J_1 \in\left(\begin{array}{c}
{[m]} \\
k_i
\end{array}\right)}\operatorname{det}((\mathbf{H}_i)_{J_1}).
\end{align}
If a sub-matrix $(\mathbf{H}_i)_{\mathcal{C}}$ are allowed to transmit with a constraint  $|\mathcal{C}|=r_0$, to minimize the difference between the left term and right term of (a) in Eq. (\ref{eq:ensure}), we can immediately obtain a sub-optimal solution by DPP greedy search as $\mathcal{C}^* = \text{MAP-DPP}(\mathbf{H}_i,r_0)$, since it has to be selected at least the first $r_0-k_i+1$ largest $\operatorname{det}((\mathbf{H}_i)_{J_1})$ in the greedy search. Here $\mathcal{C}^*$ denotes the index set of selected representative dimensions.
We define tolerable sparsity $Rm$ as the number of elements that can be losslessly transmitted from the center to each source. Transmitting the symmetric matrix $(\mathbf{H}_i)_{\mathcal{C}^*}$ requires $(r_0^2+r_0)/2$ elements, which corresponds to the number of elements in the lower triangular matrix of $(\mathbf{H}_i)_{\mathcal{C}^*}$. Furthermore, in cases where additional sparsity can be utilized for compression purposes, we can compress the residual matrix $(\mathbf{H}_i)_{\bar{\mathcal{C}}^*}$ through singular value decomposition (SVD). Here $\Bar{{\mathcal{C}}}^*=[m]\setminus \mathcal{C}^*$. By considering only the first $r_1$ singular vectors and values, we only require a sparsity of $r_1m$ for the compression. Hence, the constraint on the tolerable sparsity $R_m$ can be expressed as $(r_0^2+r_0)/2+r_1m\leq Rm$. 
Mathematically, 
\begin{align}\label{eq:sparse}
    &\hat{\mathbf{H}}_i = (\mathbf{H}_i)_{\mathcal{C}^*} + \mathbf{V}(1:r_1,:)\text{diag}(\lambda_1,\cdots,\lambda_{r_1})\mathbf{V}^\top(1:r_1,:) \\ \nonumber
    s.t.~& (r_0^2+r_0)/2+r_1m\leq Rm, \\ \nonumber
        &\mathcal{C}^* = \text{MAP-DPP}(\mathbf{H}_i,r_0), \\ \nonumber
        &\mathbf{V}\text{diag}(\lambda_1,\cdots,\lambda_{m})\mathbf{V}^\top = \text{svd}(\mathbf{H}_i- (\mathbf{H}_i)_{\mathcal{C}^*} ).
\end{align}
Therefore, the data samples in each source can be pre-coded as $\widetilde{\mathbf{Z}}_{A_i} = \mathbf{Z}_{A_i}\hat{\mathbf{H}}^{1/2}_i$. In fact, since the CSI information is not completely reliable, we precode the data samples conservatively and use a momentum way, which is presented as $\widetilde{\mathbf{Z}}_{A_i} =\mathbf{Z}_{A_i}\mathbf{W}_i= \mathbf{Z}_{A_i}(\mathbf{I}+\hat{\mathbf{H}}^{1/2}_i)$.
A summary of our approach is shown in Algorithm 1.

\begin{algorithm}[ht]\label{alg:1}
\small
\setstretch{0.6}
\caption{ DDPP: MAP Inference for MAP for Distributed Data Source  } 
\begin{algorithmic}
\REQUIRE Source data $\mathbf{Z}_{S_1}, \cdots,\mathbf{Z}_{S_N}$, Center information $\mathcal{Y}=\{Y_1,\cdots,Y_N\}$, the number of items selected in each interval $K$. Sparsity parameters $r_0,r_1$. The index set of selection $\mathcal{A}$.
\ENSURE The index set of selection $\mathcal{A}$.
\FOR {$i$ \textbf{in} $1:N$} \item\textcolor{cyan}{\#In each source $i$.}
\textcolor{cyan}{\#All sources do this step in parallel.}
\item Computed CSI information $\hat{\mathbf{H}}_i$ based on Eq. (\ref{eq:sparse}).
\item Pre-code using: $\widetilde{\mathbf{Z}}_{A_i} = \mathbf{Z}_{A_i}(\mathbf{I}+\hat{\mathbf{H}}^{1/2}_i)$.
\item Compute pre-coded Gram matrix: $\widetilde{\mathbf{L}}_{S_i} = \widetilde{\mathbf{Z}}_{A_i}\widetilde{\mathbf{Z}}^\top_{A_i}$.
\item $A_i\leftarrow\emptyset$
  \WHILE{$|A_i|\leq k_i$}
\item $j=\arg \max _{i \in S_i \backslash {A_i}} \log \operatorname{det}\left(\widetilde{\mathbf{L}}_{A_i \cup\{i\}}\right)-\log \operatorname{det}\left(\widetilde{\mathbf{L}}_{A_i}\right)$
\item \textbf{if} $j\notin A$ \textbf{then} $A_i\leftarrow A_i\cup j$  \textbf{end if}
\ENDWHILE
\ENDFOR
\item $\mathcal{A}\leftarrow \mathcal{A}\cup A_i\cup \cdots\cup A_N$.
\end{algorithmic}
\end{algorithm}
\vspace{-0.2in}

\begin{table}[]
\centering
\caption{Some important notations used in Sections \ref{sec:Formulation} and \ref{sec:Methodology}.}
\label{tab:notations}
\resizebox{0.47\textwidth}{!}{%
\begin{tabular}{ll} \toprule
\multicolumn{1}{c}{\textbf{Notation}} & \multicolumn{1}{c}{\textbf{Description}} \\ \midrule
$A_i$ & The index set of selected samples from source $i$. \\
$\mathcal{A}$ & \begin{tabular}[c]{@{}l@{}}The index set of selected   samples from all sources.\\      $\mathcal{A}=A_1\cup A_2\cup A_i\cup\cdots\cup A_N $.\end{tabular} \\
$\mathcal{B}$ & The index set of selected samples until a moment. \\
$\mathcal{C}$ & The indices of selected columns and rows of $\mathbf{H}_i$. \\
$k_T$ & The total number of selected samples. \\
$k_i$ & The number of selected samples from source $i$. \\
$[k_i]$ & The set $\{1,\cdots,k_i\}$. \\
$\mathbf{L}$ & The Gram matrix of the entire dataset. $\mathbf{L}=\mathbf{Z}\mathbf{Z}^\top$. \\
$\mathbf{L}_{A_i}$ & The Gram matrix of data from source $i$. $\mathbf{L}=\mathbf{Z}_{A_i}\mathbf{Z}_{A_i}^\top$. \\
$m$ & The dimensions of the dataset. \\
$[m]$ & The set $\{1,\cdots,m\}$. \\
$\left(\begin{array}{c}{[m]}   \\k_i\end{array}\right)$ & The set of $k_i$-combinations of $[m]$ \\
$n$ & The number of samples in the dataset. \\
$n_i$ & The number of samples in source $i$. \\
$N$ & The total number of sources. \\
$S_i$ & The index set of all samples from source $i$. \\
$r_0$ & The cardinality of $\mathcal{C}$,. $|\mathcal{C}|=r_0$. \\
$r_1$ & The number of transmitted singular vectors in Eq. (\ref{eq:sparse}). \\
$R$ & The total tolerable sparsity in transmission. \\
$\mathcal{S}$ & The index of entire dataset. \\
$Y_i$ & \begin{tabular}[c]{@{}l@{}}The index set of selected   samples at this moment\\       that are \textbf{not} from source   $S_i$.\\      i.e. $Y_i= \mathcal{B}\setminus A_i$.\end{tabular} \\
$\mathcal{Y}$ & $\mathcal{Y}=\{Y_1,\cdots,Y_N\}$. \\
$\mathbf{H}_i$ & MIMO-like CSI. Please refer to the text below Eq. (\ref{eq:12}). \\
$\hat{\mathbf{H}}_i$ & The approximation of $\mathbf{H}_i$. \\
$\mathbf{Z}$ & The data matrix of the dataset. \\
$\mathbf{Z}_{A_i}$ & The data matrix of the source $i$. \\
$\widetilde{\mathbf{Z}}_{A_i}$. & $\mathbf{Z}_{A_i}$ after pre-coding. \\ \midrule
$X_{A,B}$ & \begin{tabular}[c]{@{}l@{}}A  submatrix of $X$ with rows and   columns\\      indexed by $A$ and $B$, respectively.\end{tabular} \\ \midrule
$X_{A}$ & \begin{tabular}[c]{@{}l@{}}If $A=B$ and $X$ is a square,   $X_{A}$ can be denoted as $X_A$ (i.e. $\mathbf{L}_{A_i}$), \\      otherwise $X_A$ only denotes the rows of a non-squared matrix $X$\\      indexed by $A$ (i.e. $\mathbf{Z}_{A_i}$).\end{tabular} \\ \bottomrule
\end{tabular}%
}
\end{table}

\section{Experiment}
\subsection{Comparison Method}
In our experiments, we use the exact greedy search proposed in \cite{chen2018fast} across all samples as the \textbf{Ground Truth}. We consider multiple alternative methods for comparison, including
\begin{itemize}
    \item \textbf{GreeDi}: This is the most known distributed solution, in which, in the first round, each source greedily finds a set of size $\alpha k_T$ samples, and in
the second round, it performs another greedy search on all candidates $N \alpha k_T$ from the previous round.  To meet a zero-communication overhead policy, we set $\alpha =1/N$. It is also equivalent to our methods that discards the feedback mechanism. 
    \item \textbf{MaxDiv Source}: It performs the exact greedy search in one source with the largest RD-based diversity measured as $\log\operatorname{det}(\mathbf{I}+\frac{m}{|S_i|\epsilon}\mathbf{Z}_{S_i}^\top \mathbf{Z}_{S_i})$\cite{chen2023rd}.
    \item \textbf{Random Selection}: It involves randomly selecting samples from each source to send to the center (with the same total number as other methods). 
    \item \textbf{Stratified Sampling}: It involves randomly selecting the same number of samples from each source to send to the center. 
    \item \textbf{Greedymax}: In the first round, each source greedily finds a set of size $k_T$ samples, and in
the second round, the set from the source with the maximum diversity, is sent to the center.
    
\end{itemize}



\subsection{Dataset and Experiment Setup}\label{sec:setup}

The first experiment is conducted using two datasets, which are CIFAR10 \cite{krizhevsky2009learning} and CIFAR100 \cite{krizhevsky2009learning}. Image datasets were preferred due to the following reasons: i) they have relatively high raw data dimensions, which enables DPP to choose a subset with a larger number of samples, ii) low-dimensional semantic features can be easily obtained by pre-trained modes, and iii) image datasets are compatible with various potential applications such as drone-based aerial monitoring \cite{boroujeni2023ic} and AI-based traffic monitoring \cite{chen2022network}. 
As a proof-of-concept experiment, we used a pre-trained ResNet-18\footnote{The model can be found at \url{https://pytorch.org/vision/stable/models.html}. } to extract the latent features of images and set $m=512$. For the sake of completeness, we consider a different number of sources $N$ as 5, 10, 12, 15, and 20. Each source assigns 500 different samples, and samples from different sources are non-overlapping. 
We executed all of our experiments on a node of a cluster with an Intel(R) Xeon(R) Gold 6148 CPU with 125 gigabytes of memory and an NVIDIA A100 with 80 gigabytes of memory. Note that the primary algorithm does not require a GPU to operate. For simplicity, we consider selecting total $k_T=120$ samples in $t_T=2$ intervals, denoting only the feedback mechanism is performed once and each interval selection 60 samples. We use set tolerable sparsity defined in Section \ref{sec:csi} to $R=0.75\times k_T/t_T=45$. Note that $R\geq k_T/t_T$ leads to a trivial scenario where we can send all the previously received samples back to each source.

\subsection{Result}

\begin{figure*}[ht]
\centering\includegraphics[width=0.9\textwidth]{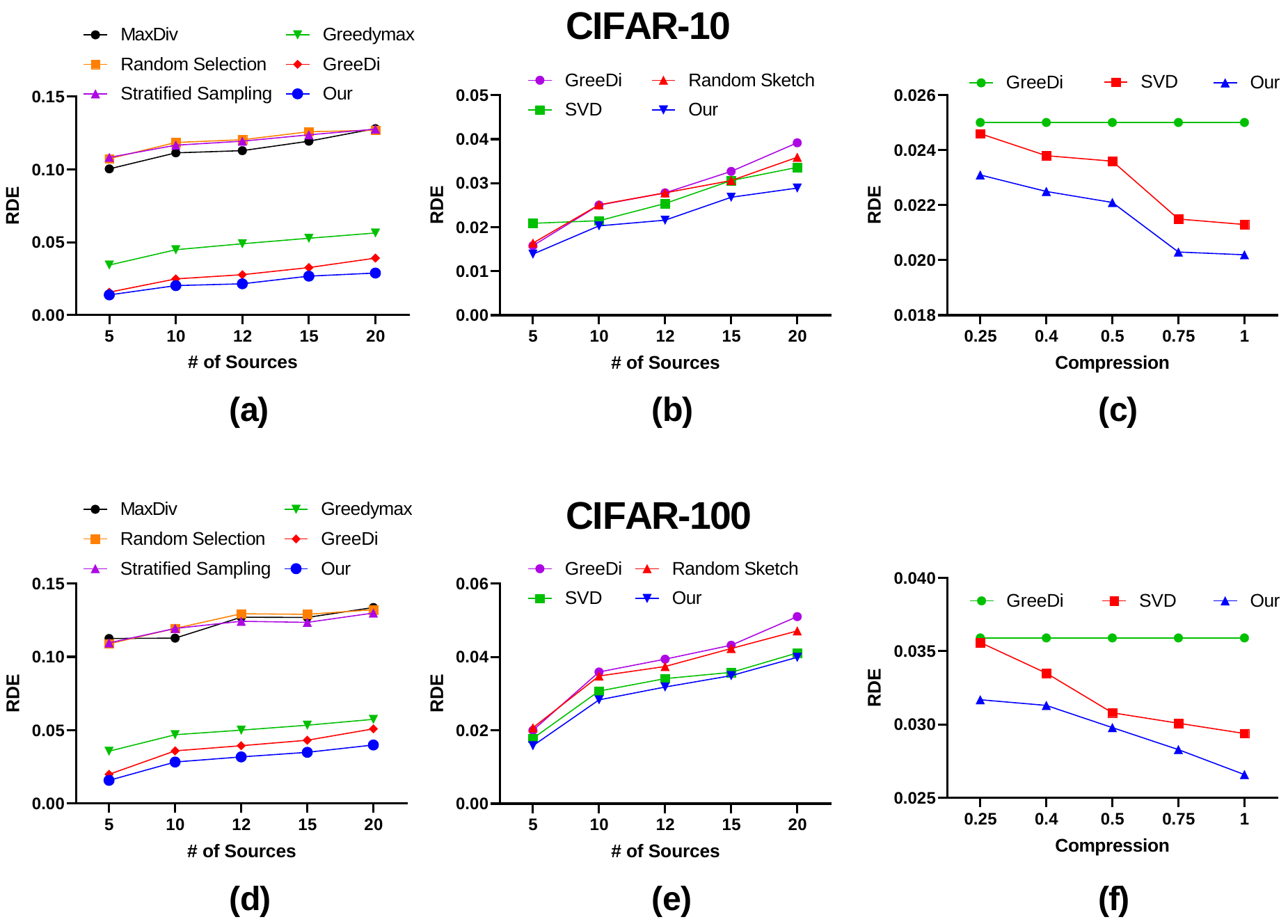}
\caption{Comparison of the performance ($\downarrow$) of different selection strategies on CIFAR-10 and CIFAR-100 datasets. } 
\label{fig:per0}
\end{figure*}
The original DPP-diversity for a selected subset $\widetilde{A}$ is defined as $\operatorname{det}(\mathbf{Z}_{\widetilde{A}} \mathbf{Z}_{\widetilde{A}}^\top )$, where a higher value represents a higher level of diversity. 
For comparison convenience, the performance is presented as the \textbf{Relative Diversity Error (RDE)} with respect to the ground truth. Specifically, if the ground truth is $A^*$ and the inference subset by one approach is $\widetilde{A}$. Then, the RDE is defined as $1-\log\operatorname{det}(\mathbf{Z}_{A^*} \mathbf{Z}_{A^*}^\top)/\log\operatorname{det}(\mathbf{Z}_{\widetilde{A}} \mathbf{Z}_{\widetilde{A}}^\top)$, with a range of 0 to 1, and lower is better. The results from running 20 times are shown in Table \ref{tab:result1} and Fig. \ref{fig:per0}(a)(d). 
Our approach outperforms all baselines and alternative methods and obtains gain for different numbers of sources on both datasets. In particular, our approach substantially improved and only has at most 25\% RDE compared to MaxDiv Source, Random Selection and Stratified Sampling. More importantly, we found the feedback mechanism can decrease RDE varying from 12\% to 26\% in CIFAR-10 and 19\% to 21\% in CIFAR-100 compared to the method without feedback. We also perform the two-sample $t$-Test to demonstrate the reliability of our method. According to the results shown in Table \ref{tab:stat}, the P-value is tiny and much smaller than 0.05, which indicates our method indeed has a statistical difference from GreeDi.

\begin{table}[ht]
\centering
\caption{The relative diversity error (RDE) of different selection methods. ($\downarrow$): Lower is better.}
\label{tab:result1}
\resizebox{0.5\textwidth}{!}{%
\begin{tabular}{llccccc}\toprule
\multicolumn{1}{l}{\textbf{Dataset}} & \textbf{\#ofCluster} & \textbf{5}      & \textbf{10}     & \textbf{12}     & \textbf{15}     & \textbf{20}     \\ \midrule
\multirow{7}{*}{\textbf{CIFAR10}}    & Ground Truth               & 0.0000          & 0.0000          & 0.0000          & 0.0000          & 0.0000          \\
                                     & MaxDiv Source               & 0.1004          & 0.1115          & 0.1130          & 0.1194          & 0.1281          \\
                                     & Random Selection     & 0.1074          & 0.1186          & 0.1205          & 0.1259          & 0.1270          \\
                                     & Stratified Sampling  & 0.1083          & 0.1167          & 0.1194          & 0.1238          & 0.1277          \\
                                     & Greedymax            & 0.0344          & 0.0450          & 0.0491          & 0.0528          & 0.0565          \\
                                     & GreeDi               & 0.0158          & 0.0250          & 0.0278          & 0.0327          & 0.0392          \\
                                     & \textbf{Our}         & \textbf{0.0139} & \textbf{0.0203} & \textbf{0.0216} & \textbf{0.0268} & \textbf{0.0209} \\ \midrule
\multirow{7}{*}{\textbf{CIFAR100}}   & Ground Truth               & 0.0000          & 0.0000          & 0.0000          & 0.0000          & 0.0000          \\
                                     & MaxDiv Source               & 0.1126          & 0.1128          & 0.1271          & 0.1270          & 0.1337          \\
                                     & Random Selection     & 0.1090          & 0.1195          & 0.1295          & 0.1291          & 0.1322          \\
                                     & Stratified Sampling  & 0.1096          & 0.1195          & 0.1243          & 0.1236          & 0.1300          \\
                                     & Greedymax            & 0.0357          & 0.0470          & 0.0501          & 0.0534          & 0.0574          \\
                                     & GreeDi               & 0.0200          & 0.0359          & 0.0394          & 0.0432          & 0.0510          \\
                                     & \textbf{Our}         & \textbf{0.0158} & \textbf{0.0283} & \textbf{0.0318} & \textbf{0.0349} & \textbf{0.0399} \\ \bottomrule
\end{tabular}%
}
\end{table}

\begin{table}[]
\centering
\caption{The P-value of two-sample t-Test between methods GreeDI and our proposed method. }
\label{tab:stat}
\resizebox{0.5\textwidth}{!}{%
\begin{tabular}{clccccc} \toprule
\multicolumn{1}{l}{\textbf{Dataset}} & \textbf{\#ofCluster} & \textbf{5} & \textbf{10} & \textbf{12} & \textbf{15} & \textbf{20} \\ \midrule
\textbf{CIFAR10}                     & GreeDI-Our           & 1.1E-04    & 1.1E-07     & 3.6E-09     & 6.3E-10     & 1.9E-16     \\
\textbf{CIFAR100}                    & GreeDI-Our           & 1.6E-07    & 2.4E-11     & 1.4E-08     & 8.0E-08     & 9.1E-11 \\ \bottomrule   
\end{tabular}%
}
\end{table}

\subsection{Ablation Analysis}
We conduct the following experiments to investigate the advantages caused by our method to compress CSI used in feedback. 
To demonstrate the performance of our compression scheme to compress the CSI $\mathbf{H_i}$, we compare the following alternative method,
\begin{itemize}
    \item \textbf{SVD} uses replaces the proposed way of compressing $\mathbf{H_i}$ with an SVD-based compression. i.e. Use first $R$ singular vector of $\mathbf{H_i}$.
    \item \textbf{Random Sketch} generate $\mathcal{C}$ by randomly sampling from $[m]$ in Eq. (\ref{eq:sparse}). As before, $[m]=\{1,\cdots,m\}$ represents the index set of dimensions. 
\end{itemize}

The results are shown in Table \ref{tab:aba1} and Fig. \ref{fig:per0}(b)(e), which confirm both SVD and the proposed method can be used to enhance the selection and our method outperform both two alternative compression strategies. We found Random Sketch, at some time, is even worse than the selection without any feedback (GreeDi), which may indicate that the actual system is vulnerable and can be easily misled by introducing undesired information.  

\begin{table}[]
\centering
\caption{Comparison of the performance ($\downarrow$) by different compression strategies with different numbers of sources. }
\label{tab:aba1}
\resizebox{0.5\textwidth}{!}{%
\begin{tabular}{llccccc}\toprule
\textbf{Dataset}                   & \textbf{\#ofCluster} & \textbf{5}      & \textbf{10}     & \textbf{12}     & \textbf{15}     & \textbf{20}     \\ \midrule
\multirow{4}{*}{\textbf{CIFAR10}}  & GreeDi               & 0.0158          & 0.0250          & 0.0278          & 0.0327          & 0.0392          \\
                                   & SVD                  & 0.0289          & 0.0215          & 0.0254          & 0.0306          & 0.0336          \\
                                   & Random Sketch        & 0.0164          & 0.0251          & 0.0278          & 0.0306          & 0.0359          \\
                                   & \textbf{Our}         & \textbf{0.0139} & \textbf{0.0203} & \textbf{0.0216} & \textbf{0.0268} & \textbf{0.0289} \\ \midrule
\multirow{4}{*}{\textbf{CIFAR100}} & GreeDi               & 0.0200          & 0.0359          & 0.0394          & 0.0432          & 0.0510          \\
                                   & SVD                  & 0.0178          & 0.0307          & 0.0341          & 0.0358          & 0.0411          \\
                                   & Random Sketch        & 0.0207          & 0.0348          & 0.0374          & 0.0423          & 0.0471          \\
                                   & \textbf{Our}         & \textbf{0.0158} & \textbf{0.0283} & \textbf{0.0318} & \textbf{0.0349} & \textbf{0.0399}\\ \bottomrule
\end{tabular}%
}
\end{table}

Additionally, we evaluate the different tolerable sparsity ($R$) as shown in Table \ref{tab:ablation_sparsity} and Fig. \ref{fig:per0}(c)(f), which show that our method can outperform SVD-based compression in different tolerable sparsity, and even at 0.25 level, our methods can decrease over 10\% RDE compare to GreeMi and SVD in CIFAR-100 dataset. 
\begin{table}[]
\centering
\caption{Comparison of the performance ($\downarrow$) by different compression strategies with different tolerable sparsity. }
\label{tab:ablation_sparsity}
\resizebox{0.5\textwidth}{!}{%
\begin{tabular}{clccccc}\toprule
\multicolumn{1}{l}{\textbf{Dataset}} & \textbf{Sparsity ($\times k_T/t_T$)} & \textbf{0.25}   & \textbf{0.4}    & \textbf{0.5}    & \textbf{0.75}   & \textbf{1}      \\ \midrule
\multirow{3}{*}{\textbf{CIFAR10}}    & GreeDi                    & 0.0250          & 0.0250          & 0.0250          & 0.0250          & 0.0250          \\
                                     & SVD                       & 0.0246          & 0.0238          & 0.0236          & 0.0215          & 0.0213          \\
                                     & \textbf{Our}              & \textbf{0.0231} & \textbf{0.0225} & \textbf{0.0221} & \textbf{0.0203} & \textbf{0.0202} \\ \midrule
\multirow{3}{*}{\textbf{CIFAR100}}   & GreeDi                    & 0.0359          & 0.0359          & 0.0359          & 0.0359          & 0.0359          \\
                                     & SVD                       & 0.0356          & 0.0335          & 0.0308          & 0.0301          & 0.0294          \\
                                     & \textbf{Our}              & \textbf{0.0317} & \textbf{0.0313} & \textbf{0.0298} & \textbf{0.0283} & \textbf{0.0266} \\ \bottomrule
\end{tabular}%
}
\end{table}

\section{Potential Applications}

In the practical data delivery system, the ultimate goal is not only selecting diversity, but also expect to improve the QoE, such as learning quality in the downstream tasks. Translating diversity gain to learning tasks (e.g., classification, object detection, etc.) often needs to adopt a proper distance. Designing a proper measure is out of our focus in this paper; however, as mentioned in Section \ref{sec:setup}, we use features extracted by a pre-trained ResNet-18 as the measure, which offers low-dimensional semantic representations that can benefit both efficient selection and enhance learning quality for image data. This idea is also employed as the \textit{knowledge base} in semantic communication networks \cite{shi2021semantic}. We show the following examples of learning tasks enhanced by the data selection with this distance measure.

\subsection{Classification}
\begin{figure*}[htbp]
\centering\includegraphics[width=0.95\textwidth]{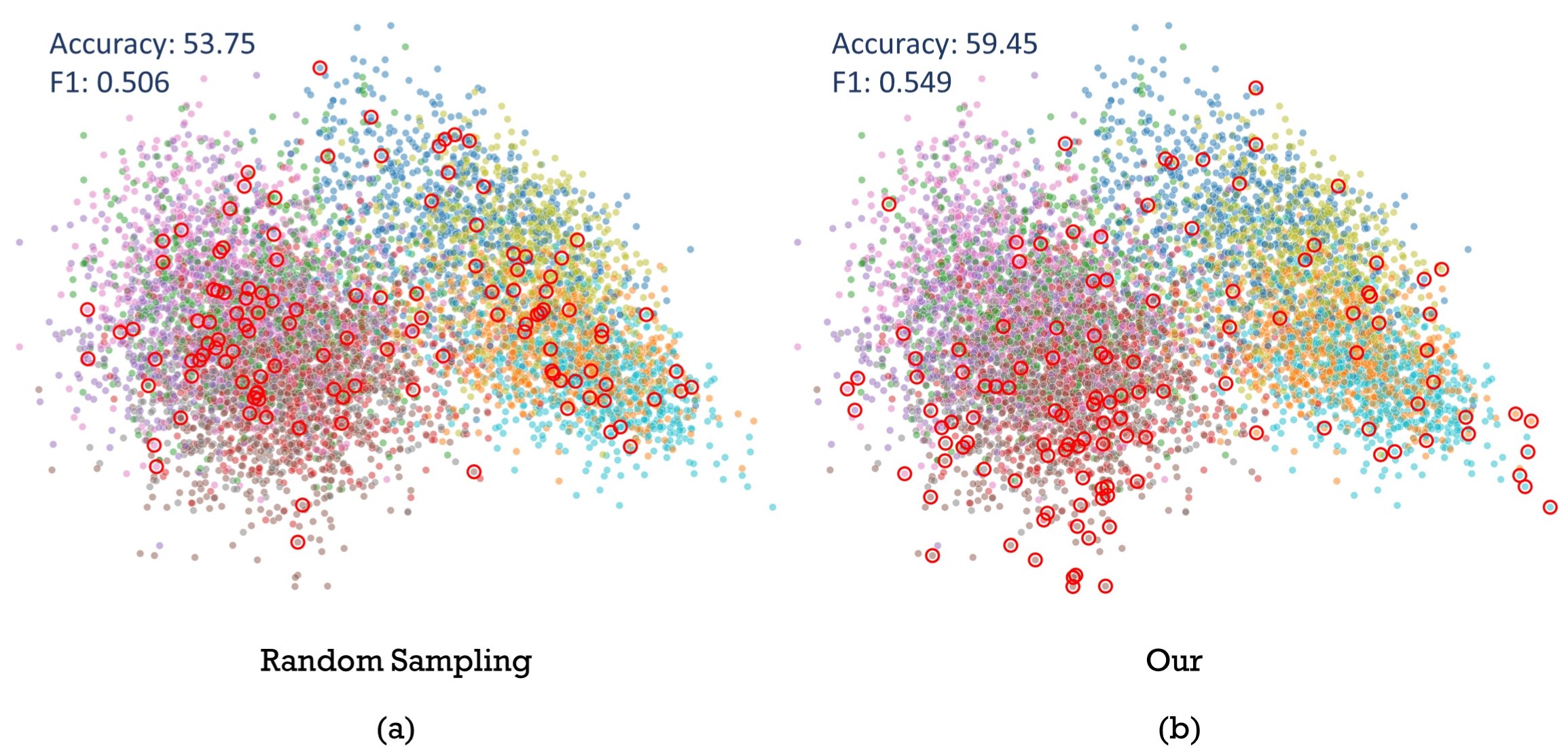}
\caption{Selected data points by different methods by PCA visualization. The different points' colors denote different data labels, and the red circles mark the selected samples.  }
\label{fig:cls0}
\end{figure*}

We first conduct the experiment of classification on CIFAR-10 and CIFAR-100 datasets. We use the k-nearest neighbors (KNN), a non-parametric method, to evaluate the representation of selected samples. Considering the different total classes of the two datasets, we set the number of nearest neighbors that are used to make a classification decision to 9 and 5, respectively. The classification results on both datasets are shown in Tables \ref{tab:CLS-1} and \ref{tab:CLS-2}, respectively, which shows learning on the data selected by our methods can outperform all other methods on both datasets. For example, our method can achieve at least 3\% accuracy and 0.02 F1 score improvement on CIFAR-10. Fig. \ref{fig:cls0} shows the selection by different methods in CIFAR-10 visualized by Principal component analysis (PCA). It is observed our approach can select more diverse data, and the diversity improvement finally is translated to the learning improvement.

\begin{table}[]
\centering
\caption{Comparison of classification results on CIFAR-10 by different strategies.}
\label{tab:CLS-1}
\resizebox{0.46\textwidth}{!}{%
\begin{tabular}{llccccc}\toprule
\textbf{Method} & \textbf{\#of Sources} & \textbf{5} & \textbf{10} & \textbf{12} & \textbf{15} & \textbf{20} \\ \midrule
\multirow{2}{*}{\textbf{Random}} & \textbf{Accuracy} & 53.16 & 52.79 & 52.66 & 52.92 & 53.96 \\
 & \textbf{F1} & 0.533 & 0.483 & 0.499 & 0.500 & 0.519 \\ \midrule
\multirow{2}{*}{\textbf{\begin{tabular}[c]{@{}l@{}}Stratified\\      Sampling\end{tabular}}} & \textbf{Accuracy} & 53.85 & 53.16 & 53.43 & 54.26 & 53.76 \\
 & \textbf{F1} & 0.517 & 0.512 & 0.509 & 0.519 & 0.514 \\ \midrule
\multirow{2}{*}{\textbf{GreeDi}} & \textbf{Accuracy} & 55.52 & 54.84 & 54.62 & 54.65 & 53.05 \\
 & \textbf{F1} & 0.523 & 0.512 & 0.502 & 0.514 & 0.500 \\ \midrule
\multirow{2}{*}{\textbf{Our}} & \textbf{Accuracy} & \textbf{58.51} & \textbf{57.27} & \textbf{57.40} & \textbf{58.05} & \textbf{56.52} \\
 & \textbf{F1} & \textbf{0.560} & \textbf{0.540} & \textbf{0.543} & \textbf{0.554} & \textbf{0.538}\\ \bottomrule
\end{tabular}%
}
\end{table}

\begin{table}[]
\centering
\caption{Comparison of classification results on CIFAR-100 by different strategies.}
\label{tab:CLS-2}
\resizebox{0.46\textwidth}{!}{%
\begin{tabular}{llccccc}\toprule
\textbf{Method} & \textbf{\#of Sources} & \textbf{5} & \textbf{10} & \textbf{12} & \textbf{15} & \textbf{20} \\ \midrule
\multirow{2}{*}{\textbf{Random}} & \textbf{Accuracy} & 46.71 & 45.58 & 44.26 & 44.15 & 42.43 \\
 & \textbf{F1} & 0.439 & 0.431 & 0.420 & 0.428 & 0.392 \\ \midrule
\multirow{2}{*}{\textbf{\begin{tabular}[c]{@{}l@{}}Stratified\\ Sampling\end{tabular}}} & \textbf{Accuracy} & 46.52 & 46.05 & 43.36 & 44.33 & 44.55 \\ 
 & \textbf{F1} & 0.442 & 0.432 & 0.406 & 0.410 & 0.418 \\ \midrule
\multirow{2}{*}{\textbf{GreeDi}} & \textbf{Accuracy} & 48.08 & 48.48 & 45.72 & 45.73 & 44.79 \\
 & \textbf{F1} & 0.441 & 0.456 & 0.414 & 0.399 & 0.391 \\ \midrule
\multirow{2}{*}{\textbf{Our}} & \textbf{Accuracy} & \textbf{49.68} & \textbf{49.02} & \textbf{47.22} & \textbf{47.78} & \textbf{46.29} \\
 & \textbf{F1} & \textbf{0.471} & \textbf{0.459} & \textbf{0.442} & \textbf{0.452} & \textbf{0.421}\\ \midrule
\end{tabular}%
}
\end{table}

\subsection{Traffic Sign Detection}

\begin{figure*}[!htbp]
\centering\includegraphics[width=0.95\textwidth]{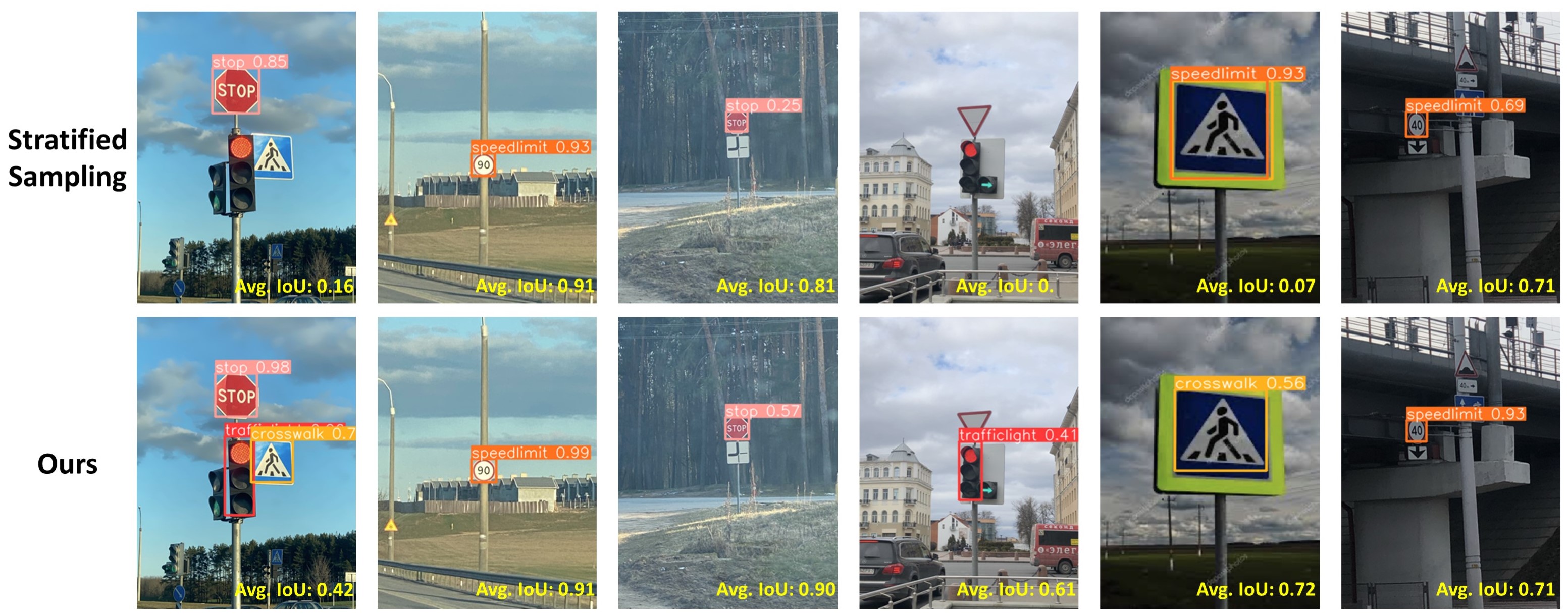}
\caption{The selected results of detection by training a YOLOv8 detector from scratch using data selected by \textbf{Top:} \textit{Stratified Sampling} and \textbf{Bottom:} Our proposed method. Our method demonstrates higher confidence and higher accuracy in the detection. } 
\label{fig:det}
\end{figure*}

Traffic sign detection is an important application in smart transportation \cite{razi2023deep}, which involves the identification and localization of traffic signs in images, typically for the purpose of assisting autonomous vehicles, advanced driver assistance systems, or traffic management. As the proof of concept, we use a small dataset \cite{signdataset} in our experiments. This dataset contains 877 images of 4 distinct classes, including Traffic Lights, Stops, Speedlimit, and Crosswalk. We use 177 images as the test set. We set the number of sources $N=10$, and each source contains 70 images. We select $k_T=120$ samples from this dataset by using different selection strategies. Then, use the selected samples to train a state-of-the-art object detector YOLOv8 \cite{Jocher_YOLO_by_Ultralytics_2023} from scratch. The performance of detection is evaluated by mean average precision (mAP) at different IoU thresholds (0.5 and 0.5-0.95) and F1 score, and we report the results over running 10 times. As shown in Table \ref{tab:object_result}, increasing the diversity generally can enhance the detection performance in the scenario. For example, with a 0.05 RDE decrease compared with Random selection, our method can obtain around 0.05 mAP improvement and 0.04 F1 score improvement, respectively. Fig. \ref{fig:det} demonstrates the selected images inference by the model trained on different selected data, which brings visible improvement in both accuracy and confidence of the detection.

\begin{table}[!]
\centering
\caption{Comparison of performance on object detection of different selection strategies.}
\label{tab:object_result}
\resizebox{0.46\textwidth}{!}{%
\begin{tabular}{clcccc}\toprule
\multicolumn{2}{c}{\textbf{\begin{tabular}[c]{@{}c@{}}Sampling   \\      Method\end{tabular}}} & \textbf{Random} & \textbf{\begin{tabular}[c]{@{}c@{}}Stratified \\      Sampling\end{tabular}} & \textbf{GreeMi} & \textbf{Our}    \\ \midrule
\multicolumn{2}{c}{\textbf{RDE} ($\downarrow$)}                                                         & 0.1077          & 0.0986                                                                       & 0.0795           & \textbf{0.0535} \\ \midrule
                                         &  \textbf{mAP@50}        & 0.5920          & 0.5923                                                                       & 0.6364            & \textbf{0.6488} \\
                                         & \textbf{mAP@50-95}    & 0.4427          & 0.4553                                                                       & 0.4734            & \textbf{0.4961} \\
\multirow{-3}{*}{\textbf{Detection}}     & \textbf{F1 score}                                   & 0.5724          & 0.5811                                                                       & 0.6016            & \textbf{0.6120}\\ \bottomrule
\end{tabular}%
}
\end{table}

\begin{table}[]
\centering
\caption{The learning quality of multiple-instance learning by different data selection strategies.}
\label{tab:ccd}
\resizebox{0.45\textwidth}{!}{%
\begin{tabular}{clllll}\toprule
\multicolumn{2}{c}{\textbf{\begin{tabular}[c]{@{}c@{}}Sampling\\      Method\end{tabular}}} & \textbf{Random} & \textbf{\begin{tabular}[c]{@{}c@{}}Stratified \\      Sampling\end{tabular}} & \textbf{GreeMi} & \textbf{Ours}   \\ \midrule
\multicolumn{2}{c}{\textbf{RDE} ($\downarrow$)}                                                            & 0.1495                      & 0.1467                         & 0.0960            & 0.0939          \\ \midrule
\multirow{3}{*}{\textbf{MIL}}                      & \textbf{Accuracy}                      & 0.8251                      & 0.8247                         & 0.8359            & \textbf{0.8548} \\
                                                   & \textbf{F1 score}                      & 0.8295                      & 0.8321                         & 0.8385            & \textbf{0.8549} \\
                                                   & \textbf{AUC}                           & 0.9066                      & 0.9038                         & 0.9094            & \textbf{0.9228}\\ \bottomrule
\end{tabular}%
}
\end{table}

\begin{figure*}[!htbp]
\centering\includegraphics[width=0.95\textwidth]{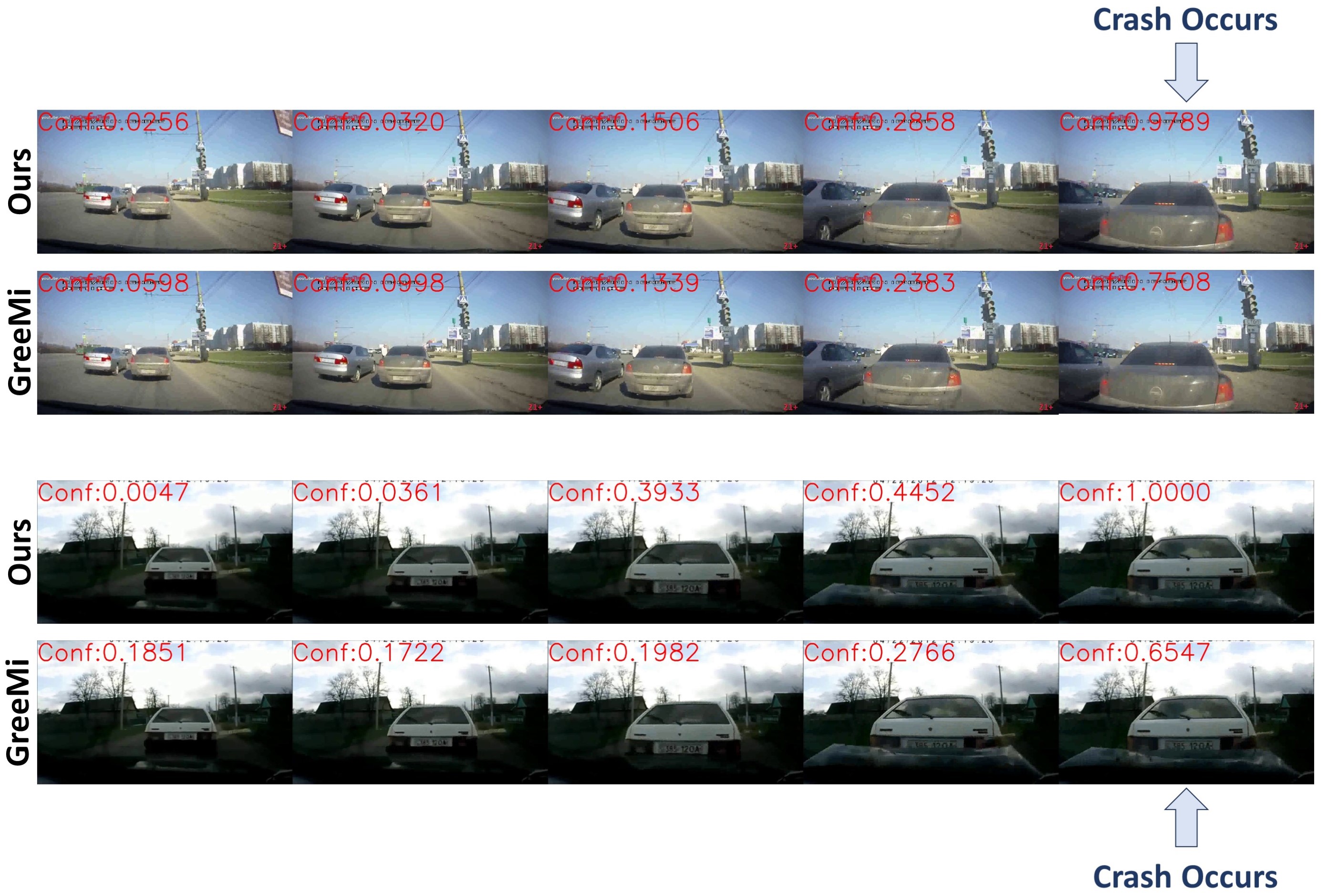}
\caption{The confidence of the crash occurs provided by the attention map in a sequential frames by two selection methods.} 
\label{fig:ccd}
\end{figure*}

\subsection{Car Crash Detection via Multiple-instance Learning}

In real surveillance videos, the task of classifying whether a crash has occurred is of paramount importance, but what is even more critical is the rapid detection of the precise moment when a crash occurs. However, this presents a significant challenge as annotating each individual frame in the video is a time-consuming and labor-intensive process, made even more complex by the varying lengths of different videos. To address this issue, a practical solution is to only leverage video-level annotation and employ Multiple Instance Learning (MIL) problem \cite{ilse2018attention}, which is a weakly-supervised method and widely used in Anomaly Detection \cite{gong2022multi,zhu2023pdl,zhu2023self,lv2023unbiased}.

In this problem, each frame in a surveillance video is treated as an instance within a bag and is assigned a binary label, either 0 for negative or 1 for positive, based on the presence or absence of a crash event. The video, on the other hand, serves as a bag containing multiple instances, which are the frames. A key characteristic is that a bag is considered negative (labeled as 0) only if all its instances, meaning all the frames within the video, are negative. Otherwise, if any frame within the video is positive, the entire bag is labeled as positive (1), indicating the presence of a crash event. In this work, an attention-based Multiple Instance Learning (MIL) method \cite{ilse2018attention} is utilized. This method likely involves the use of attention mechanisms to weigh the importance of different instances (frames) within a bag (video) when making the final classification decision, effectively allowing the model to focus on the most informative frames for crash detection. 

We use the Car Crash Dataset (CCD) \cite{BaoMM2020}, which collected traffic accident videos captured by dashcams mounted on driving vehicles. We choose $N=10$ and each source contains 100 videos. We select $k_T=80$ videos using $t_T=2$ intervals.
The diversity and learning quality of selected samples is shown in Table \ref{tab:ccd}, which demonstrates selecting by our proposed approach, accuracy, F1 score, and AUC can be improved over all other methods. The frame-level prediction confidence of the crash obtained by the attention map from the trained model is shown in Fig. \ref{fig:ccd}, which demonstrates in both exemplary frame sequences, the model trained by samples selected by our method has the more accurate localization of the crash occurs.

\section{Discussion}
One limitation of the presented work comes from the low-rank problem of DPP, which constrains the number of selections $k_T$ by $k_T\leq rank(\mathbf{L})\approx \min(n,m)$. While our proposed design aims to select sufficient samples from distributed sources for serving the downstream tasks, it may be constrained by the inherent limitations of DPP. Extending the dimensions of data by some techniques, such as using different feature extractors, is possible, but it could introduce extra overhead. As a pioneering solution for future work, we suggest designing a kernel that satisfies various downstream tasks.

\section{Conclusion}
DPP is a formal method to enhance data diversity for learning-based systems. However, it requires access to the entire dataset in one place, which limits its applicability to diverse sources in real-world applications. To address this key challenge, we implemented a DPP MAP inference for distributed data from multiple sources as a universal diversity-maximizing data-sharing strategy for distributed sources that only requires a lightweight feedback channel from the center to the sources with no cross-source communication requirement. To this end, a novel scheduling policy, inspired by MIMO systems, is proposed. Specifically, we demonstrated that the lower bound of the original diversity maximization problem that maximizes global diversity can be decomposed into a
sum of MIMO-like individual problems. Additionally, approximating the lower bound to the original problem can be treated as receiving \textit{CSI} and \textit{pre-coding}. Under communication bandwidth constraints, we derive a sparse CSI representation to preserve the determinant via the Cauchy–Binet formula. Our experiments demonstrate that our scalable approach can compete with all alternative methods in various datasets. Moreover, as a proof-of-concept, we show that with proper distance measures, pursuing diversity can translate into improving learning quality in multiple applications, including multi-level classification, object detection, and multiple-instance learning. We expect our approach can substantially influence the design of future AI-based networking platforms, which require efficient processing of large-scale data from distributed sources.


\bibliographystyle{IEEEtran}  
\bibliography{references}




\vfill

\end{document}